\documentclass[letterpaper, 10 pt, conference]{ieeeconf}  %

\IEEEoverridecommandlockouts                              %

\usepackage{amsmath} %
\usepackage{amssymb} %
\usepackage{hyperref}
\usepackage{graphicx}
\usepackage{caption}
\usepackage{multirow}
\usepackage{pifont}

\makeatletter
\let\NAT@parse\undefined
\makeatother
\usepackage[numbers,sort&compress]{natbib}

\title{\LARGE \bf
ForceMimic: Force-Centric Imitation Learning with Force-Motion Capture System for Contact-Rich Manipulation
}

\author{Wenhai Liu$^{*}$, Junbo Wang$^{*}$, Yiming Wang$^{*}$, Weiming Wang and Cewu Lu$^{\dagger}$ \\
Shanghai Jiao Tong University \\
{\tt\small \{sjtu-wenhai, sjtuwjb3589635689, sommerfeld, wangweiming, lucewu\}@sjtu.edu.cn} \\
{\small ($^{*}$ Equal contribution, $^{\dagger}$ Corresponding author)} %
}

\begin{document}

\let\oldtwocolumn\twocolumn
\renewcommand\twocolumn[1][]{%
    \oldtwocolumn[{#1}{
    \vspace{-10mm}
    \begin{center}
           \includegraphics[width=\textwidth]{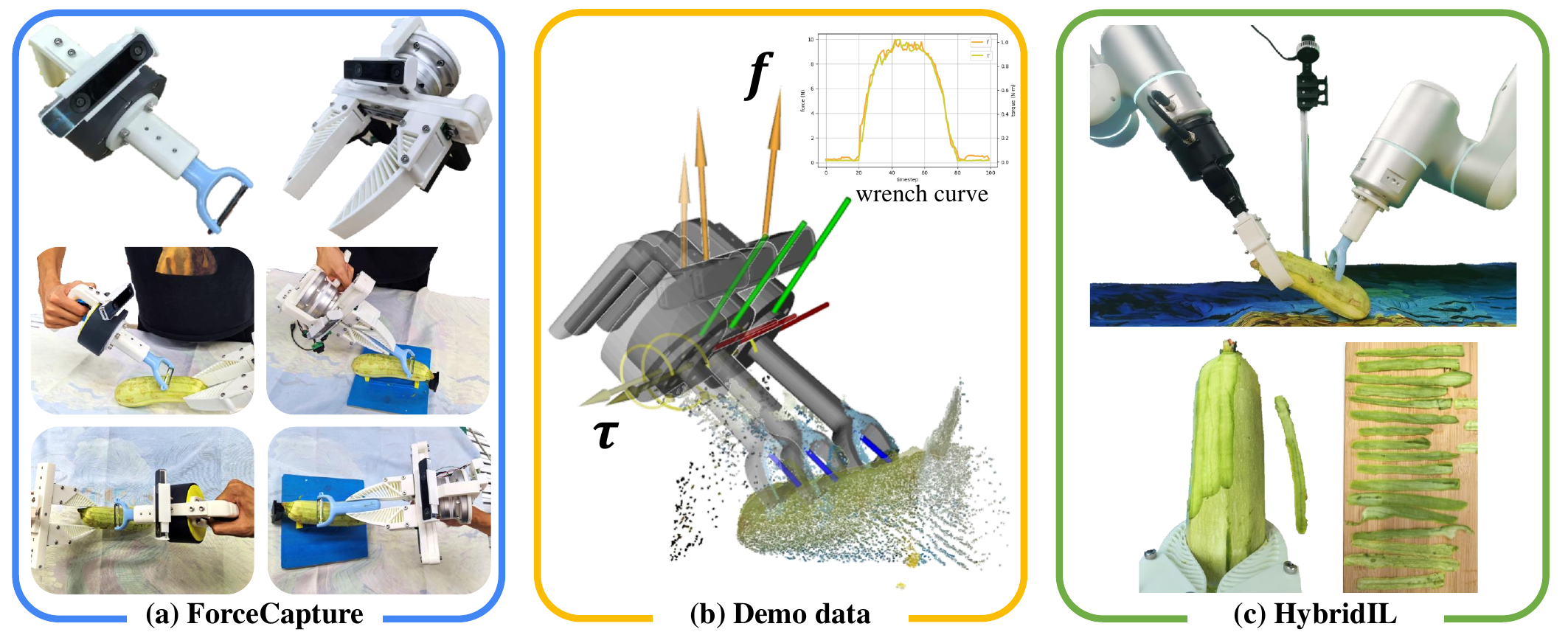}
           \captionof{figure}{Overview of \textbf{ForceMimic}. \textbf{ForceMimic} facilitates (a) \textbf{ForceCapture}, a handheld robot-free on-site data collection system, to record (b) high-quality natural force-centric manipulation demonstrations. Leveraging this data, (c) \textbf{HybridIL} trains a hybrid force-position control policy to perform contact-rich tasks, such as vegetable peeling.}
           \label{fig:teaser}
    \end{center}
    }]
}

\maketitle
\thispagestyle{empty}
\pagestyle{empty}

\begin{abstract}

In most contact-rich manipulation tasks, humans apply time-varying forces to the target object, compensating for inaccuracies in the vision-guided hand trajectory. However, current robot learning algorithms primarily focus on trajectory-based policy, with limited attention given to learning force-related skills. To address this limitation, we introduce \textbf{ForceMimic}, a force-centric robot learning system, providing a natural, force-aware and robot-free robotic demonstration collection system, along with a hybrid force-motion imitation learning algorithm for robust contact-rich manipulation. Using the proposed \textbf{ForceCapture} system, an operator can peel a zucchini in 5 minutes, while force-feedback teleoperation takes over 13 minutes and struggles with task completion. With the collected data, we propose \textbf{HybridIL} to train a force-centric imitation learning model, equipped with hybrid force-position control primitive to fit the predicted wrench-position parameters during robot execution. Experiments demonstrate that our approach enables the model to learn a more robust policy under the contact-rich task of vegetable peeling, increasing the success rates by 54.5\% relatively compared to state-of-the-art pure-vision-based imitation learning. Hardware, code, data and more results can be found on the project website at \url{https://forcemimic.github.io}.

\end{abstract}

\section{Introduction}

Humans can take use of force sensing, fine muscle force control to achieve better manipulations, from grasping~\cite{graspforce}, lifting~\cite{liftforce} to peeling~\cite{peelforce}. The exploitation of force can detect and correct the error brought by vision-based motion planning.
Inspired by these neuroscience results, we want to explore the utility of force in robot learning. However, the force-centric manipulation demonstration data is hard to collect. Substantial human videos exist on the Internet, but no interaction force data is recorded. Teleoperation~\cite{roboturk} is a popular data collection approach, enabling operators remotely to control robot finishing manipulation tasks. Particularly, force-feedback teleoperation shows a potential path to the force-centric data collection. But it cannot give the operator a natural manipulation experience, harmful to smooth action execution and precise force control. Recently, portable handheld devices~\cite{umi, dexcap} make in-the-wild learning possible. They make use of SLAM tracking camera to record human hand or handheld gripper pose trajectory. In addition to the removal of real robot, it provides an additional advantage as almost direct interaction between human and objects, which is critical in contact-rich force-centric manipulation.

On the other hand, robotic imitation learning with force involved is under-explored. Imitation policy learning mimics the function of human cerebellum, and it has been found that the central nervous system can predict the force load and even fuse this dynamics information into the inner model of human motor~\cite{graspforce}. So we wonder whether the introduction of force can help the model learn better and guide the low-level robotic control.

To handle the above challenges, we propose \textbf{ForceMimic}, a force-centric robot learning system, providing natural, force-aware and robot-free robotic demonstration collection experience and force-centric imitation learning algorithm equipped with hybrid force-position control to achieve robust contact-rich manipulation (see Fig.~\ref{fig:teaser}). We first develop \textbf{ForceCapture}, a handheld robot-free data collection system to record high-quality pure interaction wrench, i.e. combination of force and torque, by ratchet locking and gravity compensation. After that, \textbf{HybridIL} leverages the data to train a force-aware policy that outputs wrench-position parameters. And we apply hybrid force-position control to fit not only predicted pose trajectory but also the predicted force parameters, achieving robust execution against error-prone visual guide. Using ForceCapture, operators can peel a zucchini in just 5 minutes, while force-feedback teleoperation takes over 13 minutes and struggles with smooth peeling. Robot experiments show that HybridIL achieves a 54.5\% higher success rate during contact-rich peeling compared to state-of-the-art vision-based imitation learning.

Overall, our contributions can be summarized as follows:
\begin{itemize}
    \item We develop ForceCapture, a handheld robot-free data collection system, providing natural, force-aware and on-site force realism collecting experience.
    \item We provide the HybridIL algorithm, a force-centric imitation learning model that outputs wrench-position parameters and utilizes orthogonal hybrid force-position control primitives to fit the model’s predictions.
    \item We conduct experiments on robot zucchini peeling and achieve more robust performance using our data and model than state-of-the-art pure-vision-based imitation learning algorithms.
\end{itemize}

\section{Related Work}

\paragraph{Robotic Data Collection System}

A direct approach to collect robot manipulation demonstrations is teleoperation~\cite{roboturk}, where a human operator remotely controls the robot to execute the manipulation task, by various user interfaces, including haptic devices~\cite{rh20t}, exoskeleton~\cite{forceexo, airexo, ace}, virtual reality~\cite{vr, open-teach, bunny-visionpro, open-television} and leader-follower paradigm~\cite{master2robot, gello, aloha, aloha2, mobile-aloha}. Teleoperation can gather real robot data, with no domain gap between training and rollout data, but it poses the unintuitive controlling nature between human operators and robots, even when added force feedback. Recently, hand-held grippers~\cite{graspinwild, easy, umi, umi-on-legs, dexcap} make in-the-wild learning possible. However, although the hand-held gripper offers a more natural experience during data collection, it does not make the policy model aware of this interaction, with no interaction force recorded.

\paragraph{Robot Imitation Learning}

Imitation learning (IL) from human expert collected demonstrations has been widely applied in robot learning tasks. Behavior cloning (BC)~\cite{alvinn}, as one of the simplest methods in IL, directly learns the policy mapping from observations to corresponding robot actions in a supervised manner. Despite its simplicity, BC has shown many exciting results in various robot manipulations. Most methods parameterize the policy using neural networks~\cite{bet, ibc, aloha}, mapping 2D raw image pixels to the action space, while some non-parametric approaches~\cite{vinn} leverages the nearest neighbor to retrieve actions from the demonstration dataset. Recently, Diffusion Policy~\cite{diffusion-policy-rss} conditions on the vision representations and uses diffusion model to denoise the action trajectory. Built upon it, several approaches~\cite{dp3, rise} have been adapted to 3D point clouds as observation. However, current imitation learning approaches focus predominantly on trajectory-based skills, lacking exploration of action spaces such as interaction forces.

Force perception and control plays a crucial role in manipulation tasks, providing valuable and complementary information with visual guidance~\cite{sensory}. Several works have explored the force in contact-rich robotic manipulation, ranging from opening bottle caps~\cite{feel}, assembling~\cite{assembly} to playing Jenga~\cite{see-feel-act}. Recently, MOMA-Force~\cite{moma-force} utilizes the visual representation similarity to retrieve target action and wrench from the expert database and uses a PID-based controller~\cite{impedance, admittance} to control the robot. ForceSight~\cite{forcesight} presents a transformer-based robotic planner that generates force-based objectives given a text input and an RGBD image. In this paper, We propose a new paradigm of using orthogonal hybrid force-position control primitives to fit the model's predicted continue wrench-position parameters.

\paragraph{Robot Peeling}

While peeling is an important instrumental activity of daily living (IADL), it is relatively under-explored in current robot research field. Dong et al.~\cite{rulepeel} attempts peeling five types of food by calculating the cutting plane and controlling the constant contact force along the planning trajectory, but this method depends heavily on preset assumption. MORPHeus~\cite{morpheus} introduces neural networks to release the hand-crafted perception assumption, but it separates the peeling procedure into several individual modules and preset skills, focusing on high-level skill arrangement. There also exist other works dealing with the peeling problem but using knife~\cite{grapefruit} or dexterous hand~\cite{banana}, instead of peelers in our setup. In contrast to the aforementioned methods, we approach the peeling task as a force-related skill for end-to-end learning.

\section{Method}

\begin{figure*}[htbp]
    \centering
    \includegraphics[width=0.95\textwidth]{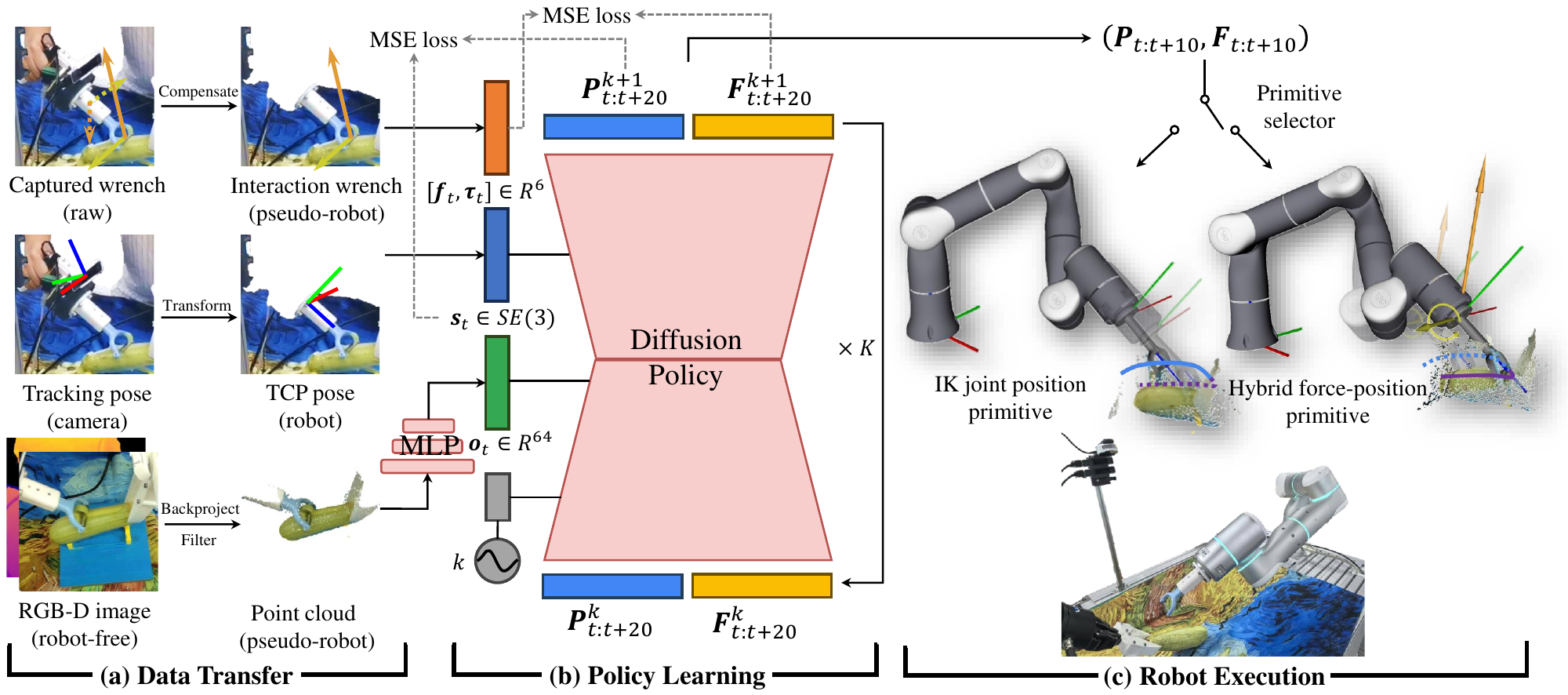}
    \caption{Overview of the pipeline. (a) We first transfer the collected robot-free data to (pseudo-)robot data, bridging the domain gap. The captured wrench is compensated to account for self-gravity effects. The pose recorded by SLAM camera is transformed as the robot TCP pose. And RGB-D observation images are backprojected into point cloud and filtered out unrelated points. (b) Leveraging this data, a diffusion-based policy is learned, with both pose and wrench predicted, conditioned on the encoded point cloud features, history pose and diffusion timestep embeddings. (c) According to the predicted force value, either IK joint position primitive or hybrid force-position primitive is selected, and fits the output force-position parameters to conduct execution actions.}
    \label{fig:pipeline}
    \vspace{-5mm}
\end{figure*}

ForceMimic first employs ForceCapture, a handheld robot-free data collection system (detailed in Sec.~\ref{sec:hardware}), to naturally gather force-centric human demonstration data. Then, we transfer the robot-free data to (pseudo-)robot data (detailed in Sec.~\ref{sec:data}), bridging the domain gap. Leveraging this data, HybridIL learns to predict wrench-pose trajectory, and applies hybrid force-position control to fit the predicted force-position parameters (detailed in Sec.~\ref{sec:software}), enabling robust performance in contact-rich manipulation tasks. The overall pipeline is illustrated in Fig.~\ref{fig:pipeline}.

\subsection{Hardware Design: ForceCapture}
\label{sec:hardware}

Accurately, naturally, and cost-effectively capturing force data during contact-rich manipulation remains a significant challenge. Inspired by existing handheld motion data collection devices~\cite{umi, dexcap}, we developed a low-cost, versatile, and robot-free force-position capture device, ForceCapture. To design ForceCapture, we consistently adhered to the following objectives:

(1) \textbf{Scalability}. Key factors for scalability include low cost, compatibility with different force sensors, ease of fabrication and maintenance.

(2) \textbf{On-site force realism}. Unlike teleoperation systems that create a sense of presence through force feedback, our goal is to directly capture real-time force data from human operations without requiring users to learn how to interact with artificial environments created by the device.

(3) \textbf{Ergonomic comfort}. The device must adhere to ergonomic principles, including an appropriate center of gravity and the convenience of operation, to ensure it does not interfere with the user's natural operating habits. Since accurate interaction force data needs to be recorded, poor ergonomics could alter muscle exertion patterns or cause discomfort, leading to non-natural force data during operation.

\begin{figure}[htbp]
    \centering
    \includegraphics[width=0.95\linewidth]{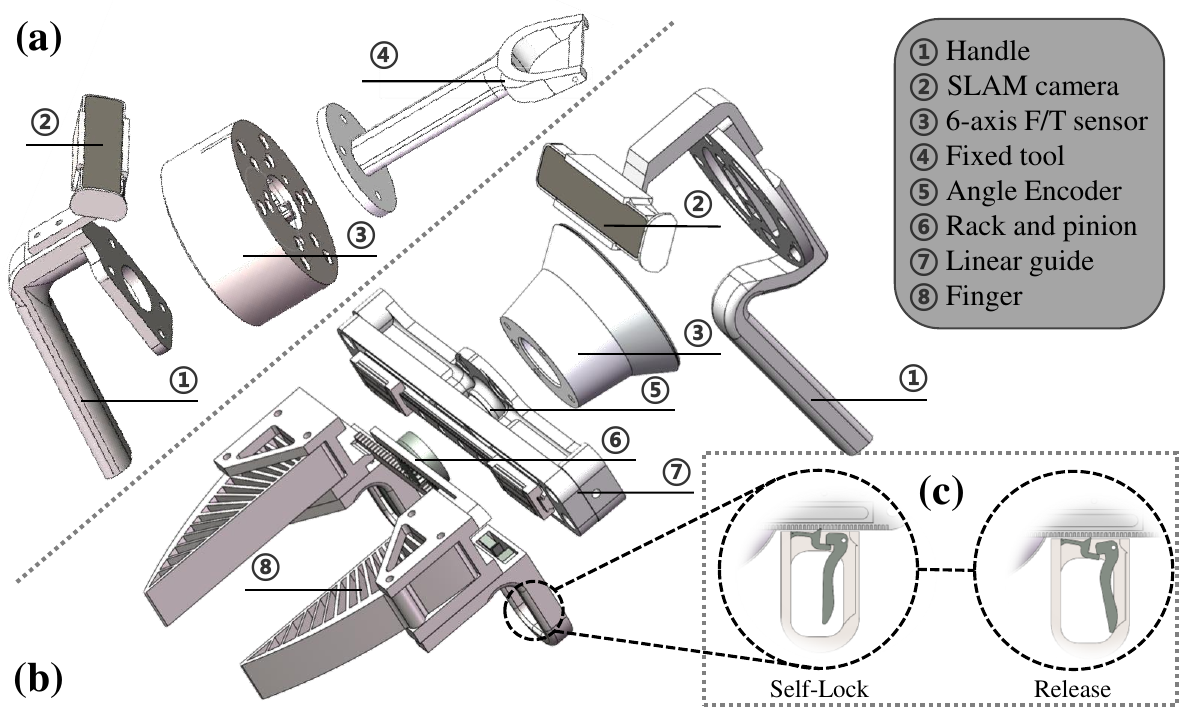}
    \caption{Structure of ForceCapture. It consists of (a) a fixed-tool end-effector version, and (b) a movable gripper version, which provides (c) a unique self-lock function.}
    \label{fig:hardware}
    \vspace{-5mm} %
\end{figure}

The overall design is shown in Fig.~\ref{fig:hardware}, which illustrates two versions, one with a fixed tool and the other with an adaptive gripper. At its core, both designs share the feature of a six-axis force sensor placed between the end-effector and the user's gripping handle, which can be used to capture the effector-environment interaction wrench. Additionally, a SLAM camera positioned near the center of the force sensor records the motion data during interaction. The user grips the handle to directly operate the tool or control the fingers for grasping and manipulation tasks. %
The rack-and-pinion mechanism of the gripper version at the base of both fingers ensures synchronized movement of the grippers. The pinion is connected to an encoder, which records the opening distance of the grippers.
The continuous width value is determined based on the calibrated relationship between the encoder angle and the gripper's width. 

It is important to note that during the manual control of the gripper’s opening and closing, the forces exerted by the hand on the grippers are also applied to the force sensor. To address this, we designed a unidirectional locking mechanism, as shown in Fig.~\ref{fig:hardware} (c). Once the fingers are closed, they cannot be opened from the fingertip. Instead, they can only be released using a lever mechanism to unlock the gripper. This design aligns with the natural logic of opening and closing the gripper and adheres to ergonomic principles. Additionally, the overall design of ForceCapture, with its center of mass positioned above the handle, conforms to the natural force application habits of the human hand.

ForceCapture is quite straightforward to manufacture, with the main body fully produced using 3D printing. The total cost of the printed parts and encoder is approximately \$50, aligning with the design goal of cost-effectiveness. The weight of the device equipped with the gripper is only 0.8kg, of which the force sensor weighs 0.5kg, and our accessories weigh only 0.3kg, which is even lighter than a can of cola. For more details about ForceCapture, including CAD models, installation instructions and 3D printing materials, please refer to the \href{https://forcemimic.github.io}{project website}.

\subsection{Data Collection and Transfer}
\label{sec:data}

The data collection system includes a six-axis F/T sensor, a RealSense T265 SLAM camera, and an external RealSense L515 RGB-D camera. For the gripper version, encoder angle data is also collected. Their respective sampling frequencies are 1000 Hz, 200 Hz, 30 Hz, and 30 Hz. Each sensor collects data at its own frequency, and during data processing, all frequencies are aligned to match the frequency of L515 observation. At the initial stage, T265 is placed on the L515 mount, and the relative position between the T265 and L515 is determined by their mounting positions. Once data collection begins, the T265 is detached from the mount and placed on ForceCapture. This process is similar to DexCap~\cite{dexcap}, where the initial position of the T265 relative to the L515 is used to track the position of ForceCapture.

ForceCapture is designed to record only interaction forces between end-effector and external environment. However, the force sensor measures the combined forces, including the tool's gravitational and inertial forces. Therefore, it is necessary to subtract the external forces generated by the tool or gripper from the force sensor data. We assume that the data collection process with ForceCapture is quasi-static, meaning that at each position, the forces are in static equilibrium, and we only need to compensate for the tool's gravity.
To perform the gravity compensation, we first move ForceCapture in a quasi-static manner for a certain period while recording the pose and wrench data. Using the static equilibrium forces at each position, we construct an overdetermined system of equations to estimate the tool's center of mass and weight using least-squares solution.

Additionally, RGB-D images recorded by L515 camera are backprojected into point clouds. To reduce discrepancies between the point clouds during data collection and those used in robot deployment, we uniformly exclude point clouds above the operational background and end-effector coordinate systems, retaining only the consistent end-effector and object point clouds. And the point clouds are voxelized to a size of 10,000. The example data transfer process is shown in Fig.~\ref{fig:pipeline} (a).

\subsection{Learning Algorithm: HybridIL}
\label{sec:software}

This section introduces HybridIL, an end-to-end imitation learning method centered on force, which maps from perception to a force-position hybrid control strategy, as shown in Fig.~\ref{fig:pipeline} (b). HybridIL takes point clouds as visual input, which are represented as one-dimensional visual features via an MLP encoder. These features are then cascaded with the robot's TCP pose to form a joint representation of multiple modalities. The strategy generation utilizes modified diffusion policy~\cite{diffusion-policy-rss} to predict both position and wrench parameters over the next 20 time steps.

\begin{figure}[htbp]
    \centering
    \includegraphics[width=0.9\linewidth]{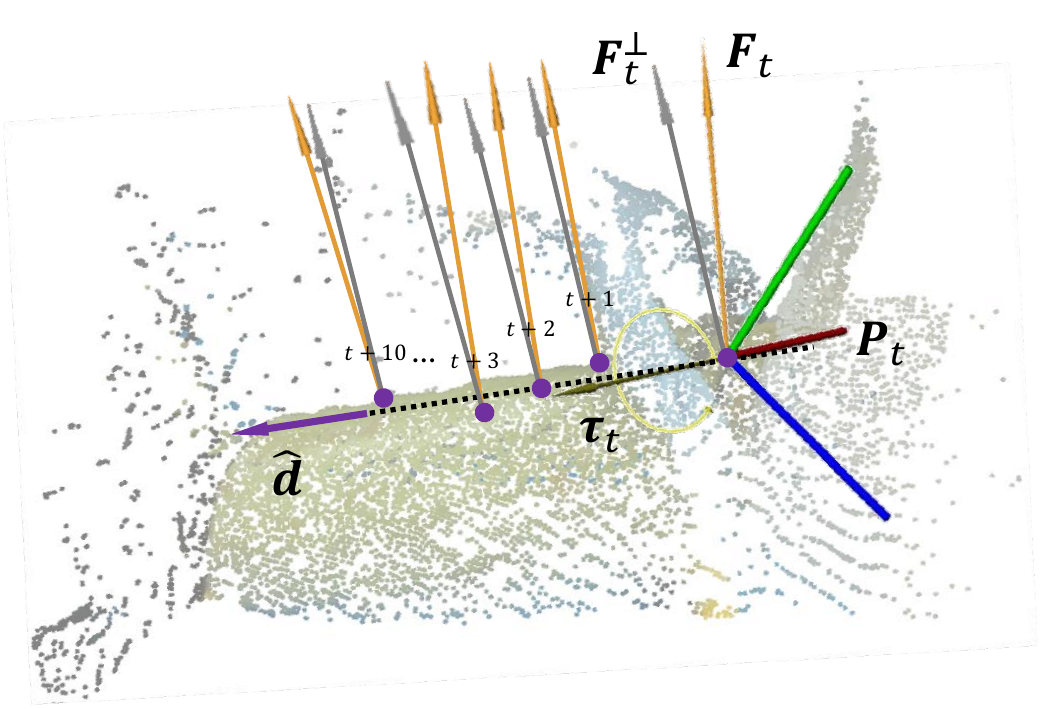}
    \caption{Illustration of the interface between policy and control primitive. When the hybrid force-position control primitive is active, the motion direction $\hat{\mathbf{d}}$ is calculated based on the pose trajectory $\mathbf{P}_{t:t+10}$ from policy, and the predicted forces $\mathbf{F}_{t:t+10}$ are orthogonalized to $\mathbf{F}^{\perp}_{t:t+10}$. Hybrid force-position control primitive then takes $\hat{\mathbf{d}}$ and $\mathbf{F}^{\perp}_t$ as parameters and controls the robot to track both pose and force.}
    \label{fig:controller}
\end{figure}

It is important to note that wrench and position control must be orthogonal. While our model does not explicitly model the orthogonality of wrench and position, we achieve this through an orthogonal force-position hybrid controller that aligns with the model's predicted force-position parameters. This approach differs from conventional imitation learning methods, which typically use a fixed lower-level position controller to track the position commands prediction by the model.
HybridIL employs two distinct control primitives to fit the model's predicted force-position parameters, demonstrated in Fig.~\ref{fig:pipeline} (c). When the predicted force is below a threshold of 6N, an IK-based~\cite{pinocchioweb} joint position controller is used. If the predicted force exceeds 6N over consecutive steps, a hybrid force-position controller is employed to execute the model's predicted parameters. The force threshold of 6N was empirically determined. The orthogonal force-position matching approach is illustrated in Fig.~\ref{fig:controller}. For force-position actions where the force exceeds 6N continuously, the motion direction is determined based on the positional information before and after. The corresponding predicted force information is projected onto the orthogonal plane of the motion direction, which defines the force control parameters during execution.
For the initial step of hybrid force-position control, if the end-effector has not yet made contact with the object, a pressing control in the opposite direction of force control is applied to achieve stable contact. These functionalities are realized using Flexiv RDK\footnote{\url{https://github.com/flexivrobotics/flexiv_rdk}} of joint position control and hybrid force-position control primitives to execute the force-position actions of HybridIL.

\section{Experiments}

In this section, we perform a zucchini peeling experiment to validate the data collection efficiency of ForceCapture and the effectiveness of HybridIL. All data were collected in an on-site manner, without any involvement of robots in the data acquisition process.

\subsection{Collection Efficiency: ForceCapture vs. Teleoperation}

\begin{figure}[htbp]
    \centering
    \includegraphics[width=0.95\linewidth]{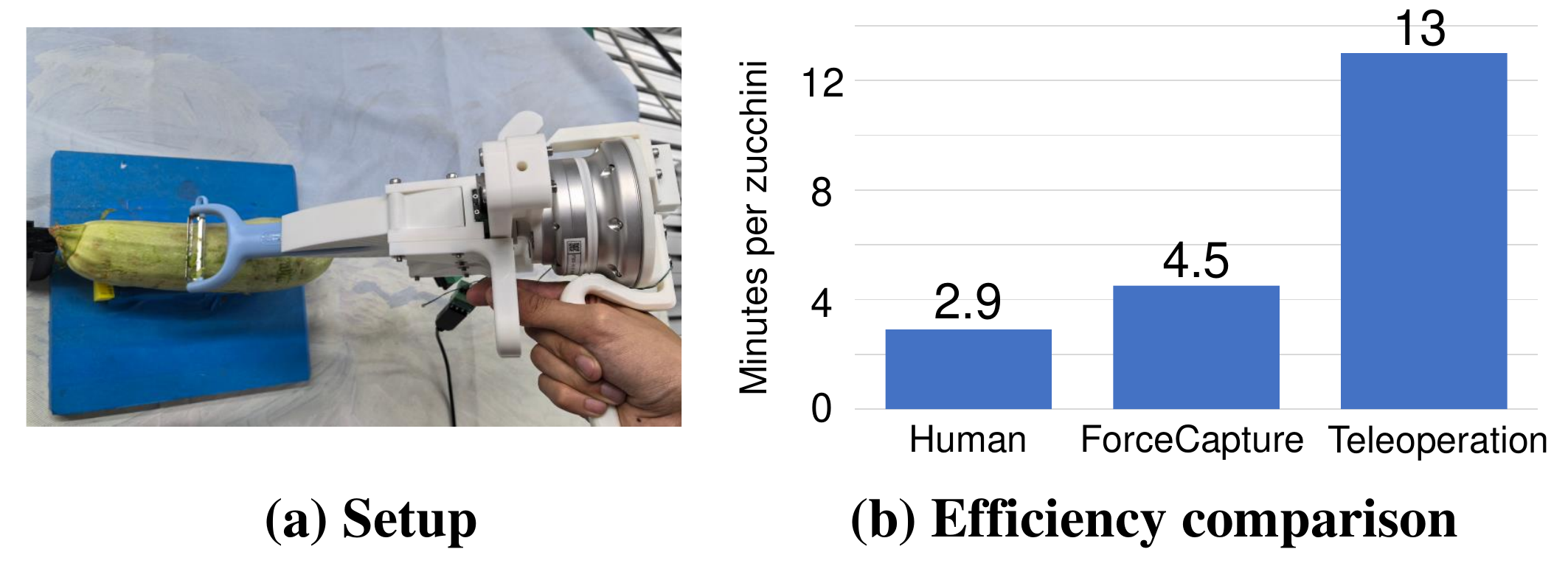}
    \caption{Experimental setup for data collection efficiency comparison and the time required to fully peel a zucchini by different methods.}
    \label{fig:efficiency-result}
\end{figure}

Currently, simultaneous collection of pose trajectory and six-axis wrench data primarily relies on teleoperation. To compare the efficiency of teleoperation with ForceCapture, we conducted a case study of peeling a zucchini using a single-arm. The experimental setup is shown in Fig.~\ref{fig:efficiency-result} (a). The procedure involved picking up the peeler, peeling the zucchini on a stand, placing the peeler down, then grasping the zucchini to adjust its orientation for peeling until the entire vegetable was peeled. Since the task involved force capture and finger movements, we used the gripper version of ForceCapture for data collection. The teleoperation setup follows the configuration described in RH20T~\cite{rh20t}.

Fig.~\ref{fig:efficiency-result} (b) shows the time comparison for completing the peeling task. The results indicate that teleoperation took approximately three times longer than ForceCapture, while the time taken by ForceCapture was very close to that of direct human peeling. It is worth noting that teleoperation requires additional training, whereas ForceCapture requires minimal training, with users becoming proficient after just one attempt. Furthermore, during teleoperation, there were three interruptions due to operational errors that caused workspace disruptions, none of which occurred with ForceCapture. ForceCapture demonstrates a more natural and streamlined data collection process, without requiring extensive user training or robotic involvement, contrasting with the more structured teleoperation setup.

\subsection{Manipulation Performance: Zucchini Peeling}

\paragraph{Setup}

To evaluate the effectiveness of ForceMimic, we formulated the peeling action as an end-to-end skill learning task. The data collection scene is exemplified in Fig.~\ref{fig:teaser} (a), utilizing the fixed-tool version of ForceCapture. The user held the zucchini steady with the left gripper and peeled with the right ForceCapture. The robot experiment setup is illustrated in the top of Fig.~\ref{fig:teaser} (c), where the L515 RGB-D camera is mounted externally to the robotic arm. The L515 camera was positioned consistently during both data collection and the robot experiment, though it can be positioned flexibly for portable in-the-wild data collection like DexCap~\cite{dexcap}. The left robot, equipped with a gripper, was used for rule-based stabilization of the zucchini, while the right arm’s fixed peeler identical to the one used in ForceCapture, performed the peeling skill via HybridIL. The robotic arm used in the experiments is the Flexiv Rizon 4, which features precise force sensing and force control capabilities.

We processed 15 zucchinis, collecting 438 peeling skill segments, resulting in a total of 30,199 action sequences. The actions advanced by 3 time steps relative to the perception data. Both the HybridIL model and the baseline methods were trained for 500 epochs each.

\paragraph{Methods}

In addition to HybridIL, we compared three other baseline methods. \textbf{Raw DP} used raw visual perception and robot pose as inputs, outputting the end-effector pose sequence based on diffusion policy. \textbf{Force DP} incorporated visual perception, robot pose, and robot force sensing as inputs, also outputting the end-effector pose sequence. \textbf{Force+Hybrid DP} used visual perception, robot pose, and robot force sensing as inputs, but output both pose and wrench sequences. For baselines that output wrench-position parameters, hybrid force-position control primitives were employed to match and switch between control modes. \textbf{Raw DP} and \textbf{HybridIL} were tested for 20 peeling actions, while other two models were tested for 10 peeling actions for their poor performance. The robot's initial TCP pose is consistent with the dataset, positioned slightly above and behind the zucchini.

\paragraph{Metrics}

We defined success using two evaluation criteria. The first criterion is whether the trajectory of the motion is correct, meaning that any length of zucchini peel is successfully removed without damaging the zucchini. The second criterion is whether a continuous peel longer than 10 cm is produced during the peeling process.

\paragraph{Results}

\begin{figure}[htbp]
\centering
\includegraphics[width=0.95\linewidth]{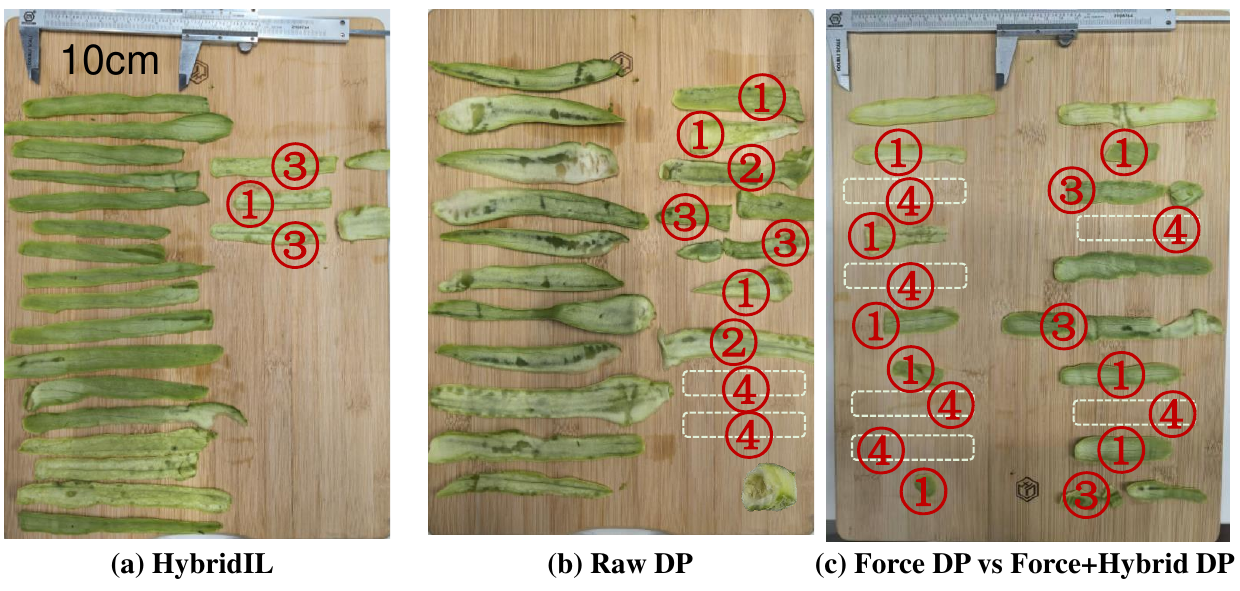} 
\caption{Visualization of the peeled skins by different methods. Failure cases are numbered with circles.}
\label{fig:qual-result}
\end{figure}

The results of the four methods are summarized in TABLE~\ref{tab:quan-result}, and the peeled skins are shown in Fig.~\ref{fig:qual-result}. The Raw DP method achieved a motion success rate of 80\% (16/20), with instances of failure detailed in Fig.~\ref{fig:qual-result} (b). Failures marked as \ding{173} involved excessive force during the peeling process, which resulted in damage to the zucchini. One instance even broke the bottom of the zucchini, as shown in the bottom of Fig.~\ref{fig:qual-result} (b). Failure marked as \ding{175} resulted from no contact with the zucchini, hence no peeling occurred. In contrast, HybridIL demonstrated a 100\% success rate (20/20), with all attempts resulting in successful contact and peeling, as illustrated in Fig.~\ref{fig:qual-result} (a). When the success criterion was increased to a continuous peeling length of more than 10 cm, both Raw DP and HybridIL experienced a decrease in success rates. Raw DP's success rate dropped to 55\%, with additional failure cases marked as \ding{172} and \ding{174}. Case \ding{172} indicated peeling lengths shorter than 10 cm, and case \ding{174} involved peeling breakage, attributed to discontinuity between the output poses, which caused peeling interruptions.
For HybridIL, the success rate decreased to 85\%, with failures in cases \ding{172} and \ding{174}. These failures were due to the premature ending of the output force-position parameters, leading to an early switch from the hybrid force control primitives to IK-based joint position control primitives, which caused peeling discontinuities.

\begin{table}[htbp]
\centering
\caption{Quantitative results of zucchini peeling.}
\label{tab:quan-result}
\begin{tabular}{ccc}
\hline
\multirow{2}{*}{\textbf{Methods}} & \multicolumn{2}{c}{\textbf{Success rate (\%)}} \\ \cline{2-3} 
                                  & motion correct                & peel length $>$ 10cm              \\ \hline
Raw DP                            & 80                     & 55                   \\
Force DP                          & 60                     & 10                    \\
Force+Hybrid DP                   & 80                     & 20                    \\
\textbf{HybridIL (proposed)}                         & \textbf{100}           & \textbf{85}          \\ \hline
\end{tabular}

\end{table}

\begin{figure}[htbp]
\centering
\includegraphics[width=0.95\linewidth]{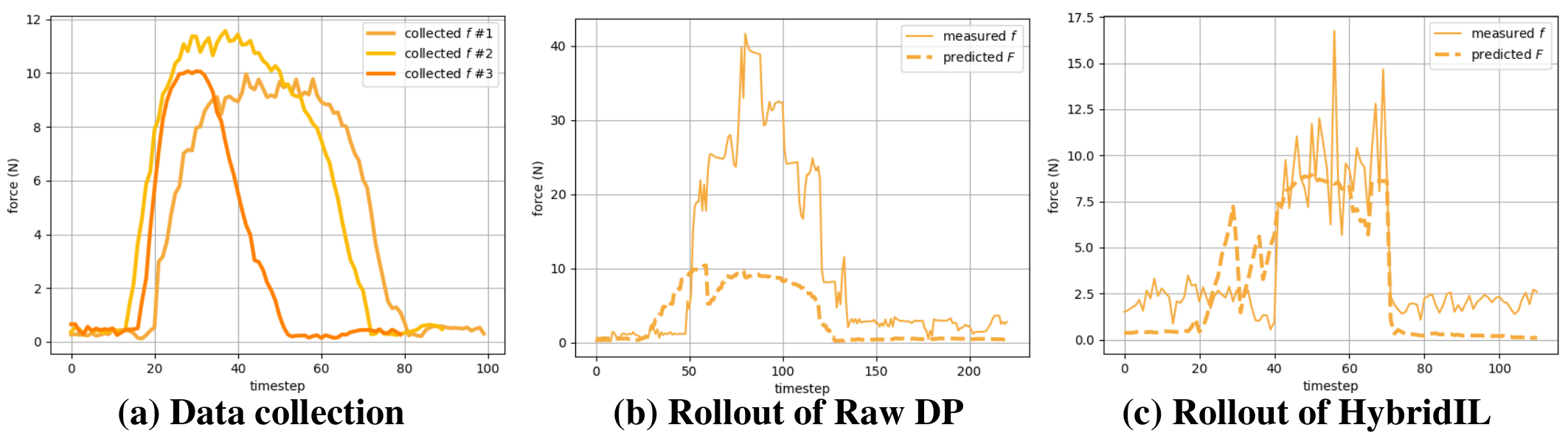} 
\caption{Examples of force curves during peeling a zucchini in different scenarios.}
\label{fig:force-result}
\end{figure}

The models that incorporate force as an input, including Force DP and Force+Hybrid DP, performed poorly. While the initial motion leading to the contact with the zucchini was generally correct, once contact occurred, these models struggled to predict the correct pose and force, making successful peeling nearly impossible. This result is counterintuitive—--one would expect that adding force sensing would improve peeling performance, but instead, it worsened the outcome.
The reasons for this can be understood from the interactive force curves of the Raw DP and HybridIL during the peeling process, as shown in Fig.~\ref{fig:force-result}. Although Raw DP successfully peeled the zucchini, the interactive forces were significantly higher, averaging around 20N and reaching over 40N in some areas. In contrast, the dataset from which these models were trained exhibited much lower interaction forces, around 10N. This mismatch between the input forces and the force distribution in the dataset made it difficult for the models to predict the correct actions. The inconsistency between the force interaction controller used during robot deployment and data collection might be a potential factor contributing to the issue. Addressing this discrepancy could improve the model’s performance. Effectively utilizing the force data collected by ForceCapture as sensory input remains an open challenge and a promising direction for future research. Further exploration on how to better integrate force perception into control strategies could lead to significant advancements in improving task performance.
Additionally, as shown in Fig.~\ref{fig:qual-result} (a) and Fig.~\ref{fig:force-result} (c), the HybridIL method maintained an average interaction force of 9N, closely aligning with the model's predicted forces, resulting in evenly peeled sections with consistent thickness and width. 
However, a slight deviation from the force distribution in the dataset was still noticeable, which likely explains why the Force+Hybrid DP failed to yield good results. Nonetheless, Force+Hybrid DP demonstrated an improvement over the Force DP.

\section{Conclusion and Discussions}

We present ForceMimic, a system aimed at advancing force-centric robot learning. This system includes ForceCapture, a scalable on-site force-position data collection system, and HybridIL, a method based on force-interaction control primitives to fit force-position parameters in imitation learning tasks. The effectiveness of both the system and method is demonstrated in the zucchini peeling task. We hope that ForceMimic will pave the way for future research on force-centric perception and hybrid force-position decision-making models in imitation learning.

We provide an initial exploration of learning human force-position skills based on ForceCapture. However, there is still room for improvement. 1) Our model uses a simple MLP to represent point clouds, robot pose, and force. In the future, we could explore more advanced multimodal representations that combine visual, force, and robot state data to improve the model's generalization to diverse skills. 2) HybridIL only employs two control primitives to fit the model's predicted force-position parameters. Future research could involve exploring more control primitives to better align with model outputs, and the model itself could potentially predict the most suitable primitive and corresponding parameters in advance. 3) ForceMimic has so far demonstrated success with a single peeling skill. In the future, the system could be extended to more force-oriented tasks.

\section*{Acknowledgements}

This work was supported by the Shanghai Commitee of Science and Technology (No. 24511103200), the National Key Research and Development Project of China (No. 2022ZD0160102), XPLORER PRIZE grants of Shanghai Artificial Intelligence Laboratory, and the National Natural Science Foundation of China (No. 52305030). We would like to thank Flexiv for the hardware of F/T sensor. We are deeply grateful to Shuhan Li, Chen Wang, Wenbo Tang, Jin Liu, Hongjie Fang, Qiaojun Yu, and Qi Wu for their invaluable insights and constructive discussions throughout this research endeavor.

{%
\bibliographystyle{IEEEtran}
\bibliography{ref}
}

\end{document}